\documentclass[USenglish,twocolumn]{article}

\usepackage[utf8]{inputenc}
\usepackage[big]{dgruyter}
\usepackage{microtype}
\usepackage{todonotes}
\usepackage{subcaption}
\usepackage{tikz}
\usepackage{pgfplots}
\usepackage{xcolor}

\begin{document}

  \articletype{...}

    \author*[1]{Felix von Haxthausen}
    \author*[1]{Jannis Hagenah}
    \author[2]{Mark Kaschwich}
    \author[2]{Markus Kleemann} 
    \author[3]{Verónica García-Vázquez} 
    \author[3]{Floris Ernst} 
        
  \runningauthor{von Haxthausen \& Hagenah et al.}
  \affil[1]{\protect\raggedright Institute for Robotics and Cognitive Systems,University of Lübeck, Lübeck, Germany. \mbox{\{vonhaxthausen, hagenah\}@rob.uni-luebeck.de}
  
  Felix von Haxthausen and Jannis Hagenah contributed equally}
  \affil[2]{Division of Vascular- and Endovascular Surgery, Department of Surgery, University Hospital Schleswig-Holstein,Lübeck, Germany}
    \affil[3]{Institute for Robotics and Cognitive Systems,University of Lübeck, Lübeck, Germany}
  \title{Robotized Ultrasound Imaging of the Peripheral Arteries - a Phantom Study}
  \runningtitle{Robotized US Imaging of the Peripheral Arteries - a Phantom Study}
  \abstract{
  The first choice in diagnostic imaging for patients suffering from peripheral arterial disease is 2D ultrasound (US). However, for a proper imaging process, a skilled and experienced sonographer is required. Additionally, it is a highly user-dependent operation. A robotized US system that autonomously scans the peripheral arteries has the potential to overcome these limitations. In this work, we extend a previously proposed system by a hierarchical image analysis pipeline based on convolutional neural networks in order to control the robot. The system was evaluated by checking its feasibility to keep the vessel lumen of a leg phantom within the US image while scanning along the artery. In 100\,$\%$ of the images acquired during the scan process the whole vessel lumen was visible. While defining an insensitivity margin of $2.74\,\mathrm{mm}$, the mean absolute distance between vessel center and the horizontal image center line was $2.47\,\mathrm{mm}$ and $3.90\,\mathrm{mm}$ for an easy and complex scenario, respectively. In conclusion, this system presents the basis for fully automatized peripheral artery imaging in humans using a radiation-free approach.
  }
  \received{...}
  \accepted{...}
  \journalname{...}
  \journalyear{...}
  \journalvolume{..}
  \journalissue{..}
  \startpage{1}
  \aop
  \DOI{...}

\maketitle

\section{Introduction} 
Today, endovascular procedures are standard in the therapy of peripheral arterial disease (PAD) \cite{one} which have increased significantly in recent years \cite{increaseprevalence}. However, a currently unsolved problem of this method is the necessity to use x-ray and contrast agent. The establishment of new radiation-free navigation methods for endovascular interventions is therefore of great relevance. In this context, ultrasound (US) as a radiation-free, affordable and real-time imaging technique has proven to be a potential alternative \cite{ASCHER20061230} .

Nevertheless, this imaging technique has several drawbacks. First, an experienced and trained sonographer is needed to avoid artifacts due to improper scanning techniques \cite{ArtifactsInUS}. Second, the inward pressure applied by the US transducer on the patient influences the anatomy to be imaged \cite{PressureAndUS}. Last, the imaging process is still highly user-dependent and time-consuming. Robotized US imaging has the potential to overcome these drawbacks.

To address the issues of an increasing number of PAD patients and US imaging disadvantages, we have previously proposed a robotic US system for semi-automatic scanning of peripheral arteries \cite{automed2020}. However, the vessel detection relied on the manual selection of a template after placing the probe. This approach assumed that the center of the vessel is the center of the template, therefore a precise selection was important.
In this work, the image analysis part used to determine if a vessel exists and to find the center of the vessel within the US image is replaced by a deep learning approach, taking the next step towards autonomy. The goal of this study was to prove feasibility, namely that the proposed system is able to continuously show the vessel lumen within a scanning process and to verify this by measuring the distance from the vessel center in the US image to the horizontal image center line.

\section{Material and Methods} 

\subsection{System Description} 

\begin{figure}[t]
\centering
	\includegraphics[width=0.8\columnwidth]{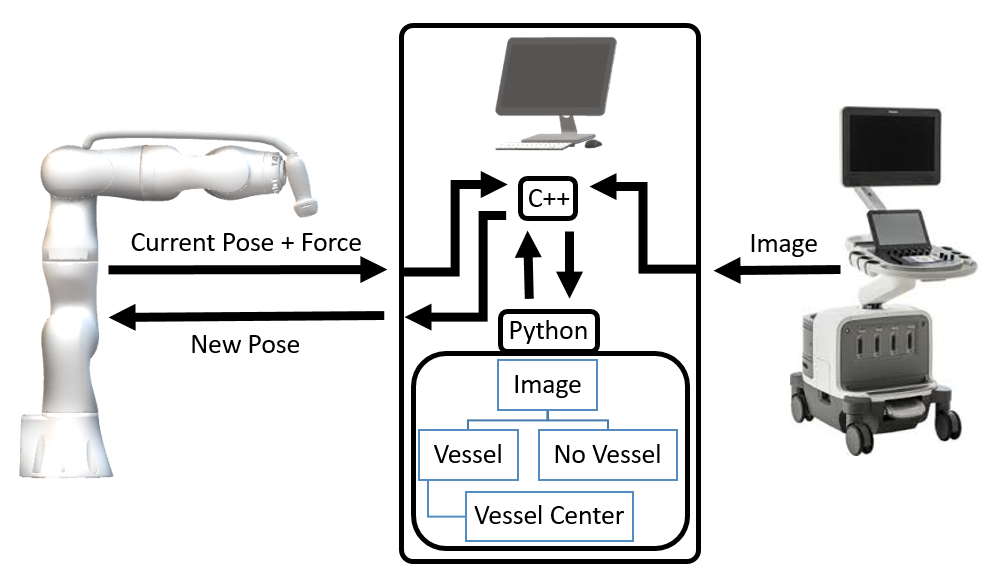}
	\caption{The system setup with its components (robotic arm, computer, US station) and the flow of data. A hierarchical image analysis takes place within python. }
	\label{img:systemsetup}
\end{figure}

A 2D linear US probe (L12-3, Philips Healthcare, Best, Netherlands) was attached to the end effector of a robotic arm (LBR iiwa 14 R820, KUKA, Augsburg, Germany) using a  custom-made  probe  holder. 8-bit grayscale images were transferred in real-time from the US station (EPIQ7, Philips Healthcare, Best, Netherlands) to the computer using a proprietary network protocol provided by Philips. An in-house middleware allows for a bidirectional communication between the robot controller and the computer. Both, the US station and the robotic arm, communicate with a C++ program running on the computer. The US images are forwarded from the C++ to a python program which in turn performs the image analysis. The result, namely the information if and where a vessel exists, is send back to the C++ program as the information is used for the robot control. The whole setup is shown in Figure \ref{img:systemsetup}.

\subsection{Image Analysis}
\label{subsec:imaganal}

The image analysis has two tasks: image classification and vessel center detection. The classification task focuses on classifying whether the vessel is visible in the image or not. If the image shows a vessel, the vessel center detection task identifies the vessel's center point in the image, leading to a regression task. Both tasks are carried out by convolutional neural networks (CNNs) with architectures roughly similar to related applications of landmark detection in medical images \cite{tetteh2018deepvesselnet,noothout2018cnn}. However, to improve the system's robustness, we decoupled both tasks and propose a hierarchical image analysis pipeline as shown in Figure \ref{img:systemsetup}. 

The classification network consists of two convolutional blocks, where each block has two convolutional layers (16 filters, size $3 \times 3$, ReLU activation) and a maximum pooling layer (size $2 \times 2$). The convolutional blocks are followed by two fully connected layers (100 neurons each, ReLU activation) and an output layer (softmax activation).
The architecture of the vessel center detection network is similar, but features three convolutional blocks with an increasing number of filters from block to block (8, 12, 16) and linear activation in the output layer to provide regression.

We acquired 8,314 US images of a leg phantom built by the Division of Vascular and Endovascular Surgery (University Hospital Schleswig-Holstein, Lübeck) in cooperation with HumanX GmbH. The same US system and probe settings were used for all acquisitions and experiments (matrix size: $277 \times 512$ pixels, image spacing: $(0.14 \times 0.14) \, \mathrm{mm^2}$, presetting: \textit{Arterial Vessel}, gain: $0 \, \mathrm{dB}$, dynamic range: 51, streaming rate: $3.9 \, \mathrm{Hz}$). The images were labeled and annotated manually. To this
end, a human observer decided for each image whether a vessel is visible ($45.9 \, \%$) or not ($54.1 \, \%$), and if yes, the vessel centerpoint was annotated. We implemented the neural networks using \textit{Tensorflow 2.0} and trained them on the given data.

\subsection{Robot Control}
\begin{figure*}[th]
    \captionsetup{justification=centering}
    \begin{subfigure}{1\columnwidth}
\begin{center}
      \includegraphics[width=.7\linewidth]{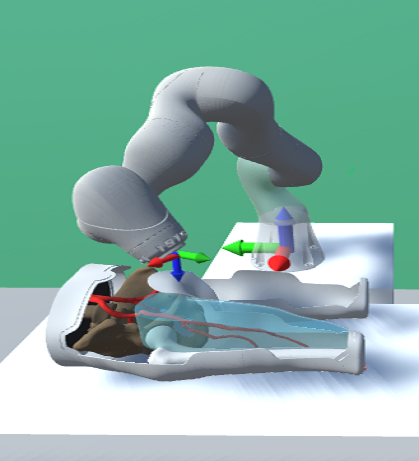}
      \caption{}
      \label{fig:virtualsetup_a}
      \end{center}
    \end{subfigure}%
    \begin{subfigure}{1\columnwidth}
\begin{center}
      \includegraphics[width=.7\linewidth]{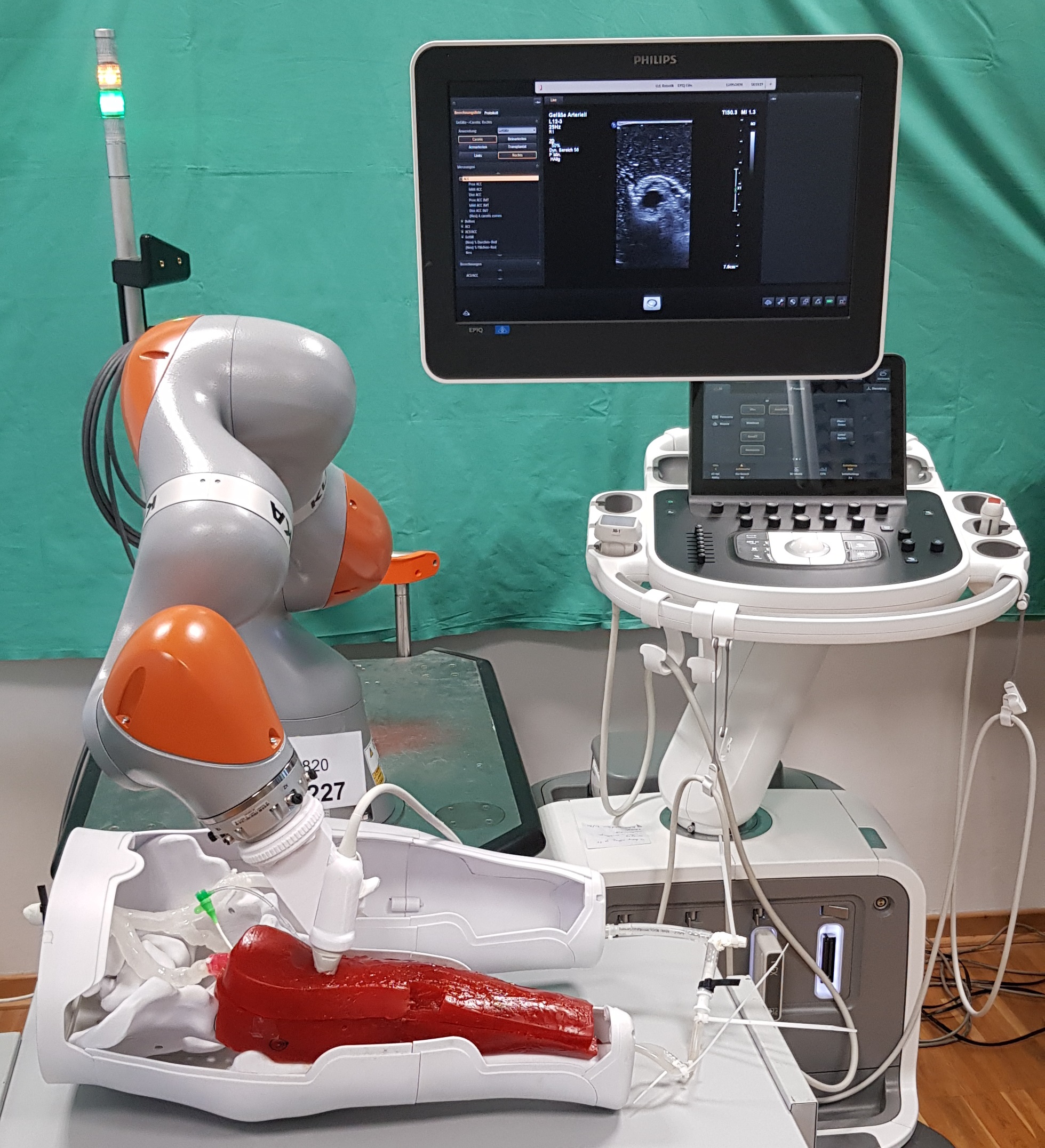}
        \caption{}
      \label{fig:virtualsetup_b}
            \end{center}
    \end{subfigure}
       
\captionsetup{justification=raggedright,singlelinecheck=false}
    \caption{(a) Virtual system setup with the robot and the leg phantom. The world coordinate system ($x$-axis - red, $y$-axis - green, $z$-axis - blue) is located at the base of the robot while the object coordinate system is at the end effector of the robot. (b) Real system setup including the US station, US probe holder and US probe.}

    \label{fig:virtualsetup}

\end{figure*}

As in our previous work \cite{automed2020}, a hand guidance mode allows the physician to place the US probe attached to the robotic arm on the area of interest. The probe is placed such that a cross-section of the vessel is visible. At this point, the automated part begins and the robotic arm is set to a proprietary cartesian impedance control mode. Each control step is based on the object coordinate system of the end effector (see Figure \ref{fig:virtualsetup_a}) and consists of the following steps:\\
The $z$-axis of the end effector is always the negative $z$-axis of the world coordinate system and corresponds to an intracorporal direction as it is assumed that the leg lays roughly parallel to the ground. In each step, the end effector is moved in positive $z$-direction until a total force of $6 \, \mathrm{N}$ is reached to keep approximately the same pressure on the leg. The $y$-axis is approximately orthogonal to the cross-section plane of the artery since the physician places it accordingly. If the classification network identifies a vessel within the US image, the robot moves $2 \, \mathrm{mm}$ in positive $y$-direction (respectively distal direction). \r{magenta}{Otherwise, the robot stops moving and the probe can be replaced.} The movement in $x$-direction aims to keep the detected vessel center in the center of the image. Thus, the information of the vessel center detection network is used to move the robot in the according opposite $x$-direction in a negative feedback loop manner. A insensitivity margin of 20 pixels ($2,74 \, \mathrm{mm}$) is implemented. This means that within this distance from the horizontal image center to the detected vessel center, no adjustment for the robot movement in $x$-direction is carried out. 

\subsection{Evaluation} 

\textbf{Image Analysis}\\
To assess the generalization, we performed a 10-fold Monte Carlo Cross Validation ($80 \, \%$ Training, $20 \, \%$ Test) for both networks (100 epochs, batchsize 64). However, for deployment in the robotic system, the models were trained on the full data set without a test split.\\

\textbf{Robot Control}\\
Two scans of the leg phantom were performed over a distance of at least $14 \, \mathrm{cm}$. The image acquisition parameters were identical to the ones mentioned in \ref{subsec:imaganal}. During the second scan, the leg phantom was turned by approximately 30° around the $z$-axis of the object coordinate system. This was to check feasibility even when the probe is not carefully placed by the physician. The quantitative evaluation regarding the robot control was twofold. First, the percentage of images acquired during the scan showing the whole vessel lumen was calculated. This is important as the lumen is used for diagnostic purposes of PAD. Second, the distance of the  horizontal image center to the true vessel center along $x$ was calculated for the saved images. This allows conclusions about the robot control and its ability to keep the vessel within the center of the image. Both metrics were assessed by manually labeling the acquired images as described in \ref{subsec:imaganal}.

\section{Results} 
\subsection{Image Analysis}

Table \ref{tab:results_cv} shows the results of the cross validations for both networks. The classification reaches an accuracy of close to $100 \, \%$, while the vessel detection network approximates the $x$-position of the vessel center with a mean absolute error of $0.47 \pm 0.36\, \mathrm{mm}$ and a maximum of $3.07 \, \mathrm{mm}$. The prediction error for the $y$-position, i.e. the vessel depth, is slightly higher.

\subsection{Robot Control}
\begin{figure}[t]
        \input{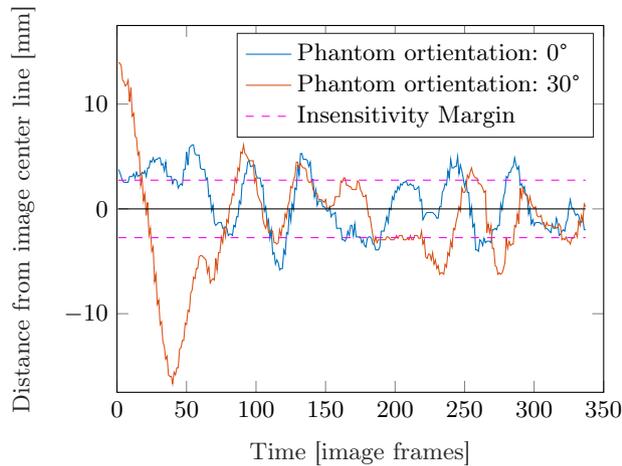}
	\caption{The distance of the vessel center in the image from the image center line in $x$-direction over time for two scans along the phantom leg (blue and red) and also the insensitivity margin (magenta).}
	\label{img:resultfigure}
\end{figure}

Both scans were successfully completed after scanning the full distance of $14 \, \mathrm{cm}$. In $100 \, \mathrm{\%}$ of the images saved during the scan, the complete vessel lumen was visible. Figure \ref{img:resultfigure} shows the distances of the  horizontal image center to the true vessel center along $x$ (MAE $2.47 \, \mathrm{mm}$ and $3.90 \, \mathrm{mm}$ for 0° and 30° respectively). Whenever the insensitivity margin is exceeded, the robot moves in the opposite $x$-direction and thus the distance from the image center line in $x$-direction decreases subsequently. However, a delay of several frames exists before the distance gradually decreases.

\begin{table}[h]
\centering
\begin{tabular}{ c c c }
 Classification & $\mu \pm \sigma$ & $99.55 \pm 0.0018$ \\
 $\mathrm{(Accuracy} \ [\%] \mathrm{)}$ & & \\
 \hline
 Vessel Detection & $\mu_x \pm \sigma_x$ & $0.47 \pm 0.36$ \\ 
 $\mathrm{(MAE} \ [\mathrm{mm}] \mathrm{)}$ & $\mu_y \pm \sigma_y$ & $0.75 \pm 0.47$ \\ 
  & $\max_x$ & $3.07$ \\   
  & $\max_y$ & $5.91$ \\   
\end{tabular}
\caption{Results of 10-fold Cross Validation for both networks. The results for the classification network are given as the relative number of rightly classified images, the vessel detection results are given as the mean absolute error (MAE) between the predicted vessel center and the ground truth.}  
\label{tab:results_cv}
\end{table}

\section{Discussion \& Conclusion}
Our results show that the image analysis works robustly on the given data. Of course, a deeper analysis of the architecture and the generalization performance has to be carried out on a real human data set. In future work, we will focus on collecting an in vivo data set from human legs as well as retraining and testing the proposed system on it.

The robotic US system is able to scan the leg phantom while keeping the vessel lumen visible within the US image. Additionally, even if the physician does not properly place the US probe on the leg phantom (non-orthogonal to the cross-section of the vessel) the system can compensate for this influence. However, the robot control only uses translational movements to follow the vessel. The system could improve the imaging by rotational adjustments due to the cylindrical anatomical shape of legs. 
Our system takes into account the total force at the end-effector. Ultimately, the pressure to be determined is the inward pressure. Therefore, future work will focus on a registration between the end effector and the US probe in order to calculate this value.
To the best of our knowledge, this is the first robotic system to automatically acquire US images of the peripheral arteries using a deep learning approach. This phantom study provides promising results and presents the basis for fully automatized peripheral artery imaging in humans using a radiation-free approach.

\begin{acknowledgement}
The authors would like to thank Till Aust for his help collecting the data set and Sven Böttger for the helpful discussions.
\end{acknowledgement}
\subsection*{Author Statement}
Research funding: This  study  was  partially  supported  by  the  German Federal Ministry of Education and Research (grant number 13GW0228) and the  Ministry  of  Economic  Affairs,  Employment,  Transport  and Technology of Schleswig-Holstein. Conflict of interest: Authors state no conflict of interest. Informed consent \& Ethical approval: Not applicable. 
\bibliographystyle{vancouver}
\bibliography{article}
\end{document}